# Distributed Pharaoh System for Network Routing

Camelia-M. Pintea[a], D. Dumitrescu[a]

[a]*Department of Computer Science, Babes-Bolyai University, Cluj-Napoca 400084, Romania*

**Abstract**

In this paper it is introduced a biobjective ant algorithm for constructing low cost routing networks. The new algorithm is called the *Distributed Pharaoh System (DPS). DPS* is based on AntNet algorithm [2]. The algorithm is using Pharaoh Ant System (PAS)[8] with an extra-exploration phase and a 'no-entry' condition in order to improve the solutions for the Low Cost Network Routing problem. Additionally it is used a cost model for overlay network construction that includes network traffic demands [1]. The Pharaoh ants (Monomorium pharaonis) includes negative pheromones with signals concentrated at decision points where trails fork [10]. The negative pheromones may complement positive pheromone or could help ants to escape from an unnecessarily long route to food that is being reinforced by attractive signals. Numerical experiments were made for a random 10-node network. The average node degree of the network tested was 4.0. The results are encouraging. The algorithm converges to the shortest path while converging on a low cost overlay routing network topology.

## 1. Introduction

Ant algorithms [5] take inspiration from the behavior of real ant colonies to identify shortest paths between the nest and a food source. While walking between their ant nest and the food source, the ants deposit a substance called pheromone. When ants arrive to a path intersection, they need to choose the path to follow. They select it applying a probabilistic decision biased by the amount of pheromone: stronger pheromone trails are preferred. The most promising paths receive a greater pheromone after some time. This is due to the fact that, because these paths are shorter, the ants following them are able to reach the goal quicker and to start the trip back soon. Finally, the pheromone is evaporated by the environment, and makes less promising paths lose pheromone because they are progressively visited by fewer ants.

In Resnick paper [9] is shown a foraging model that uses two kinds of pheromones: ants deposit one pheromone to mark trails to the food sources, but a second type of pheromone is released by the nest itself, diffusing in the environment and creating a gradient that the ants can follow to locate the nest. Resnick's ants learn to forage from the closest food source. When food sources deplete, the ants learn to search for other more distant food sources and establish paths connecting them to the nest, forgetting about the previously established trails to depleted sites.

In [8] are used three colonies of ants in order to obtain better performances. The first two colonies are explorers and the third one has exploiting ants.

The ants from the second colony are endowed with 'negative' pheromone. They return information about the possible solutions. The exploitation colony performs also exploration of the search space, using the 'no-entry' constraint and exploiting the information provided by explorer colonies.

In this work are created overlay routing networks using an ant algorithm inspired by the *Pharaoh Ant System* [8] and the AntNet algorithm described in [2]. The objectives of ant algorithm are: the shortest path between source and destination nodes while constructing a low cost overlay routing network. This biobjective ant algorithm is distributed and adaptive. While each individual node is selfish [3] in the construction of a low cost overlay, the emergent behavior of ant algorithms results in a cooperative system. Numerical experiments are made on a random 10-node network with an average node degree of 4.0. The results are graphically represented.

## 2. Prerequisites

The *Pharaoh Ant System* is based on scientific biological experiments [10]. First scope of the experiment (see Figure 1) was to see if foragers lay a negative signal on the unrewarding branch of a trail bifurcation.

Ants U-turned less often on sections from the rewarding trail; feeder branch close to the bifurcation and feeder-branch end. In the same experiment was also determined whether foragers are able to detect the negative signal before reaching the substrate on which it had been laid, zigzagging versus walking straight.

*Email addresses:* `cmpintea@cs.ubbcluj.ro` (Camelia-M. Pintea), `ddumitr@cs.ubbcluj.ro` (D. Dumitrescu)

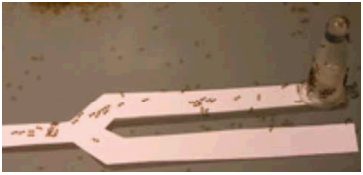

Figure 1: Pharaohs ants on a laboratory trail. One branch leads to a syrup feeder. On the other branch, a single ant heads back to the branch point, possibly laying down a stop sign pheromone.(Photo:E.J.H. Robinson [10])

The results show that significantly more ants zigzagged when approaching substrate from an unrewarding branch just after the bifurcation or at the branch end than did controls.

Fewer zigzagged when approaching substrate leading to the feeder. The results show that Pharaohs ants use a sophisticated trail system with a negative, repellent pheromone to mark unrewarding branches. The signal is concentrated at decision points trail bifurcations [7]. The pheromone being volatile provides advance warning like human road signs situated before junctions. Across a trail network, the pheromone could help direct foragers to food by closing off unrewarding sections.

It is not known yet how negative pheromones enhance foraging efficiency in trail networks, but they might complement attractive trail pheromones [6] used by Pharaoh ants in trail choice, or they could prevent strong positive feedback by attractive pheromones from locking the system into suboptimal solutions [11].

*Pharaoh Ant Sytem* [8] is based on *Ant Colony System* [4]. Three ant colonies are involved. The exploration is conducted by two colonies of ants. The second colony has - as the first colony- the role to explore the solutions' space but also a new role on spreading the negative pheromone ($Ph^-$) on bad trails. The third colony exploits the solutions domain provided by the first and second ant colony. In the exploiting phase, the third colony use the 'no-entry' signal.

In Figure 2, are shown the successive phases of ants colonies. In the first sequence an ant colony is using the positive pheromone ($Ph^+$) to mark the trails. In the second sequence other ants endowed with negative pheromone ($Ph^-$) are marking "bad" (dashed lines) some trails.

Figure 3 considers the avoiding of the "bad" trails. Some of the ants anticipate the "bad" trails zigzagging, as in the first sequence; other ants with "slow reflexes" goes for a while on the bad "trail", making a U-turn, as in the second sequence. In [8] is a general pseudo-code for the *Pharaoh Ant System*.

The *positive-Ph-ants* and the *exploiting-ants* are considered with positive pheromone $Ph^+$. A run of the algorithm returns the shortest tour found. The algorithm uses a number of $t_{max}$ iterations, selecting the best solution found during these iterations.

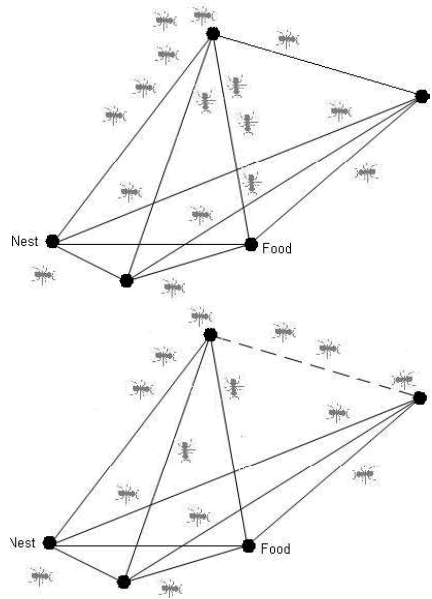

Figure 2: Successive phases of ants colonies. In the first sequence an ant colony is using the positive pheromone ($Ph^+$) to mark the trails. In the second sequence other ants endowed with negative pheromone ($Ph^-$) are marking "bad" (dashed lines) some trails.

## 3. Overlay Network Creation

Each node needs to select its neighbors in a distributed way without knowing the other node neighbors selection.

The graph representation of the overlay network is $G = (N, L)$. The physical network is represented by $G_u = (N, E)$. $N$ is the set of nodes that are in both the overlay and physical network.

The set of logical links $L$ can be different from the set of physical links $E$. A logical link $l \in L$ is setup on a path composed by physical links $e \in E$. Each node $i \in N$ has a traffic demand toward a node subset $S_i \subset N$. Let $d_{i,j}$ be the traffic demand between node $i$ and node $j$ in the subset $S_i$.

The objective for a node is to create logical links to be connected to all nodes in $S_i$ such that the total cost is minimized. The cost model defined in [1] is an extension of the cost models given in [3] to include traffic demand between nodes. The cost is defined using two cost components [1]:

- Cost to create a logical link $l_{i,j}$ from node $i$ to node $j$; the cost is proportional to the length of $l_{i,j}$ in number of hops on the physical network.

- Cost to transport the traffic demands; it is proportional to the distance between nodes $i$ and $j$ and the amount of traffic demand.

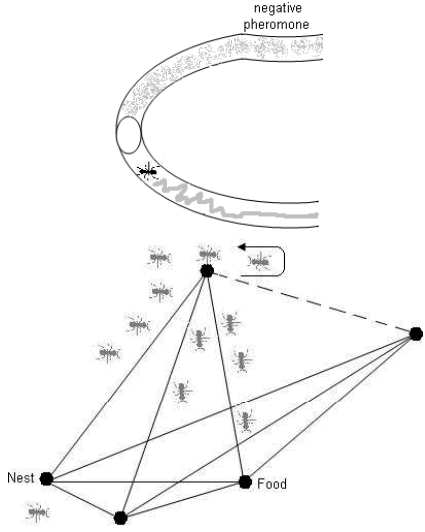

Figure 3: Avoiding the "bad" trails. Some of the ants anticipate the "bad" trails zigzagging, as in the first sequence; other ants with "slow reflexes" goes for a while on the bad "trail", making a U-turn, as in the second sequence.

As in [1] the cost for node $i$ to connect to each node $j \in S_i$ and carry traffic demand $d_{i,j}$ is defined :

$$C_i = \sum_{j \in B_i} c_h h_{i,j} + \sum_{j \in S_i} c_t t_{i,j} d_{i,j} \qquad (1)$$

where $B_i$ is the set of neighbors toward which node $i$ has a logical link $l_{i,j}$. $c_h$ is the overlay cost coefficient. $h_{i,j}$ is the number of hops on the physical network in $l_{i,j}$., $c_t$ is the underlay transit coefficient. $t_{i,j}$ is the number of transit virtual links in the path to node $j$. The total cost of the overlay network is defined as:

$$C(G) = \sum_{i \in N} C_i \qquad (2)$$

The function $C_i$ depends both on the location of the logical link $l_{i,j}$ and the demand $d_{i,j}$. Changing traffic demands in the network will cause $C(G)$ to change over time.

## 4. Distributed Pharaoh System

Based on *AntNet* [2] and *Pharaoh Ant System* [8] we introduce a new algorithm called the *Distributed Pharaoh System (DPS)*.

In [2] the forward ants are used to traverse the network. They construct a path from the source to the destination. The backward ants retrace the path updating pheromone tables and routing information.

The two objectives of the *DPS* are routing through the network and construct a low cost overlay network topology. Forward ants use the pheromone table and an estimate of the distance to the destination in determining the next node to visit.

---

**Algorithm 1** Algorithm ForwardAnt(src, dst, volume)

Initialize data structures
$path \leftarrow []$ Set the path to an empty list
$hopCount \leftarrow 0$
$current \leftarrow src$
**while** $current \neq dst$ **do**
    Pick next node
    **if** there exists an overlay $j$ for $current$ to $dst$ **then**
        $next \leftarrow j$
    **else**
        Pick a node $j$ that has not been visited, from the neighbors of the current node $i$ with a given probability
    **end if**
    Move to the next node
    Append $next$ to $path$
    $hopCount \leftarrow hopCount + hops_{current,next}$
    $current \leftarrow next$
    Update the node demand and hops from $src$
    $d_{current,dst} \leftarrow (1-\rho)d_{current,dst} + (\rho)volume$
    $hops_{current,src} \leftarrow min(hopCount, hops_{current,src})$
**end while**

---

**Algorithm 2** BackwardAnt(forwardAnt)

Initialize data structures
$src \leftarrow forwardAnt.src; \quad dst \leftarrow forwardAnt.dst$
$path \leftarrow forwardAnt.path$
$hopCount \leftarrow forwardAnt.hopCount$
$previous \leftarrow current \leftarrow dst$
$minOverlayCost \leftarrow \infty$
**while** $current \neq src$ **do**
    Update node routing parameters
    Set the number of hops from the current node to the destination node as:
    $hops_{current,dst} \leftarrow hopCount - hops_{current,src}$
    and set the desirability of the current node with regards to the destination node:
    $\eta_{current,dst} \leftarrow max(\eta_{current,dst}, \frac{1}{hops_{current,src}})$
    $\tau_{current,next}(dst) \leftarrow (1-\rho)\tau_{current,next}(dst) + \rho\tau_0$
    The cost of establishing an overlay to the current node is defined as:
    $cost \leftarrow c_h hops_{current,src} + c_t d_{current} hops_{current,dst}$
    and is used in setting the value of establishing an overlay to the current node as follows:
    $\sigma_{current} \leftarrow max(\sigma_0, (1-\rho)\sigma_{current} + \rho\frac{Q}{cost})$
    where $\rho$ is the local evaporation rate and $Q$ is a scaling parameter.
    Check if current node is best overlay node
    **if** $(cost \leq minOverlayCost)$ and $(\tau_{current,next} \geq \tau_{min})$ **then**
        $minOverlayCost \leftarrow cost; \quad overlay \leftarrow current;$
        $\sigma_{overlay} \leftarrow \sigma_{current}$
    **end if**
    Move to the next node on path
    $previous \leftarrow current; \quad current \leftarrow$ next node from $path$
**end while**

If there exists an overlay at the current node for the destination, then the ant takes the overlay link to the next node.

The new algorithm *DPS* is using three ant colonies. The first colony is using the positive pheromone ($Ph^+$) to find and mark the good trails. The second colony, as the first colony, has the role to explore the solutions space but also a new role on spreading the negative pheromone ($Ph^-$) on bad trails.

$$\tau_{i,j}(dst) \leftarrow -((1-\rho_n)\tau_{i,j}(dst) + \rho_n \tau_0) \quad (3)$$

where $\rho_n$ is the negative evaporation rate. The third colony exploits the solutions of the first two colonies. Backward ants follow the path taken by a forward ant as given in Backward AntNet [2] but also anticipating the bad trails zigzagging or making U-turns. As a consequence, a new constraint $\tau_{current,next} \geq \tau_{min}$ is introduced in order to avoid the bad trails.

At each node the backward ant calculates the cost of establishing an overlay to this node, storing the minimum cost along the path. The backward ant also updates the estimate of the number of hops remaining to the destination.

The heuristic function for each node is improved in this way iteratively. When the backward ant arrives at the source node, the source node establishes a temporary logical link to the minimum cost node along the ants path if the node does not already have a logical link established towards the destination node. Nodes are performing periodic activities such as deleting expired logical links and applying a global evaporation rule to the pheromone table:

$$\tau_{i,j}(dst) = (1-\rho_g)\tau_{i,j}(dst) \quad (4)$$

where $g$ is a global evaporation constant.

## 5. Numerical experiments

The system was implemented as specified in the previous section and tested with a randomly generated 10-node network having average node degree of 4.0. The following parameter values were used: $\alpha = 1$, $\beta = 3$, $\rho = 0.05$, $\rho_g = 0.02$, $\rho_n = 0.02$, $\tau_0 = 0.1$, $\tau_{min} = 0.01$, $Q = 100$, $\sigma_0 = 0.1$, $c_h = 2$, and $c_t = 1$. The case with a single source node and multiple destination nodes was considered. As can be seen in Figure 4 the path length of paths found from the source to a destination node converges to a minimal value. The length of the paths found to other destination nodes behaves similarly.

## 6. Conclusion

In this paper we introduced a biobjective ant colony algorithm for constructing low cost overlay routing networks. The results show the distributed nature of ant colony algorithms is ideal for quick construction of low cost overlays.

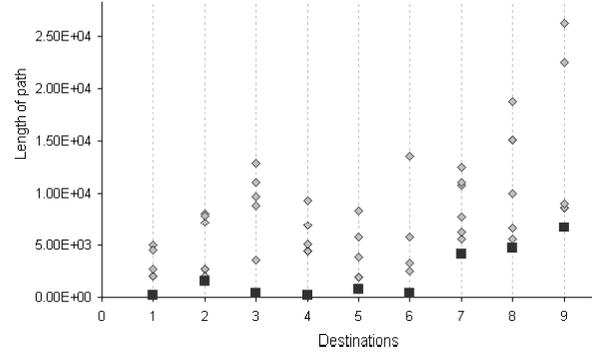

Figure 4: Path length found from a source to multiple destinations. The black point shows the minimum length for each destination.

The behavior of the ant colony algorithm makes it well suited for the dynamic behavior of traffic demands in the physical network.

This preliminary investigation is promising. Future work will include additional resource constraints. As noted in [3], the distance in the cost function (eq.1) can represent any metric such as latency or bandwidth. Other investigations will extend the algorithm to a multiobjective case considering other metrics in overlay routing network construction.

## References

[1] B. McBride, C. Scoglio, S. Das. Distributed Biobjective Ant Colony Algorithm for Low Cost Overlay Network Routing. *Proc., Int. Conf. on Artificial Intelligence*, Las Vegas, Nevada, 457–464, 2006.

[2] G.D. Caro, M. Dorigo. Antnet: Distributed stigmergetic control for communications networks. *J. of Artificial Intelligence Research*, 9:317–365, 1998.

[3] B.-G. Chun, R. Fonseca, I. Stoica, J. Kubiatowicz. Characterizing selfishly constructed overlay routing networks. *INFOCOM 2004. Twenty-third Annual Joint Conference of the IEEE Computer and Communications Societies* 2004.

[4] M. Dorigo, L.M. Gambardella. Ant Colony System: A cooperative learning approach to the traveling salesman problem. *IEEE Transactions on Systems,Man, and Cybernetics-Part B*, 26(1):29–41, 1996.

[5] M. Dorigo, G. Di Caro. The ant colony optimization metaheuristic. In D. Corne, M. Dorigo, F. Glover (Eds.), *New ideas in optimization*, London: McGraw-Hill, 11–32, 1999.

[6] M. Dorigo, T. Stützle. The ant colony optimization metaheuristic: Algorithms, applications and advances. In F. Glover and G. Kochenberger (Eds.), *Handbook of metaheuristics*, Kluwer Academic Publishers, 251–285, 2003.

[7] D.E. Jackson, M. Holcombe, F.L.W. Ratnieks. *Nature* 432:907–909, 2004.

[8] C-M. Pintea, D.Dumitrescu. Introducing Pharaoh Ant System. *MENDEL 2007, 13-th International Conference on Soft Computing, Brno Univ. of Technology. (Eds. M. Radomil and P.Osmera)*, 54–59, 2007.

[9] M. Resnick. *Turtles, Termites and Traffic Jams*. MIT Press 1994.

[10] E.J.H. Robinson, D.E. Jackson, M. Holcombe, F.L.W. Ratnieks. No entry signal in ant foraging. *Nature* 438(24):442, 2005.

[11] D.J.T. Sumpter, M. Beekman. *Anim. Behav.* 66:273–280, 2003.